\documentclass[sigconf]{acmart}

\AtBeginDocument{%
  }

\setcopyright{none} %
\copyrightyear{2026}
\acmYear{2026}
\acmDOI{}
\acmISBN{}

\acmConference[KDD AI4Mental '26]{Workshop on AI for Cognitive and Mental Health Support}{August 2026}{Jeju, South Korea}

\begin{document}

\raggedbottom

\title{Do Depressive Facial Patterns Transfer Across Cultures and Contexts? Evidence from a German RCT and E-DAIC}

\author{Misha Sadeghi}
\orcid{0000-0002-6120-0677}
\email{misha.sadeghi@fau.de}
\affiliation{%
  \institution{Dept. AIBE, FAU Erlangen-Nürnberg}
  \institution{Munich Center for Machine Learning}
  \country{Germany}
}

\author{Robert Richer}
\orcid{0000-0003-0272-5403}
\affiliation{%
  \institution{Dept. AIBE, FAU Erlangen-Nürnberg}
  \institution{Munich Center for Machine Learning}
  \country{Germany}
}

\author{Lydia Helene Rupp}
\orcid{0000-0002-4248-1055}
\affiliation{%
  \institution{Dept. of Clinical Psychology and Psychotherapy, FAU Erlangen-Nürnberg}
  \country{Germany}
}

\author{Lena Schindler-Gmelch}
\orcid{0000-0002-8355-1603}
\affiliation{%
  \institution{Dept. of Clinical Psychology and Psychotherapy, FAU Erlangen-Nürnberg}
  \country{Germany}
}

\author{Marie Keinert}
\orcid{0000-0002-7427-3058}
\affiliation{%
  \institution{Dept. of Clinical Psychology and Psychotherapy, FAU Erlangen-Nürnberg}
  \country{Germany}
}

\author{Farnaz Rahimi}
\orcid{0009-0001-0284-5221}
\affiliation{%
  \institution{Dept. AIBE, FAU Erlangen-Nürnberg}
  \country{Germany}
}

\author{Malin Hager}
\affiliation{%
  \institution{Dept. of Clinical Psychology and Psychotherapy, FAU Erlangen-Nürnberg}
  \country{Germany}
}

\author{Bernhard Egger}
\orcid{0000-0002-4736-2397}
\affiliation{%
  \institution{Dept. of Computer Science, FAU Erlangen-Nürnberg}
  \country{Germany}
}

\author{Matthias Berking}
\orcid{0000-0001-5903-4748}
\affiliation{%
  \institution{Dept. of Clinical Psychology and Psychotherapy, FAU Erlangen-Nürnberg}
  \country{Germany}
}

\author{Bjoern M. Eskofier}
\orcid{0000-0002-0417-0336}
\affiliation{%
  \institution{Dept. AIBE, FAU Erlangen-Nürnberg}
  \institution{Chair of AI-supported Therapy Decisions, LMU München}
  \institution{Helmholtz Zentrum München}
  \institution{Munich Center for Machine Learning}
  \country{Germany}
}

\renewcommand{\shortauthors}{Sadeghi et al.}

\begin{abstract}
Automated assessment of depression from facial dynamics holds promise for scalable mental health monitoring, yet cross-corpus generalization of learned biomarkers remains an open challenge. We present a systematic bidirectional transfer study pairing the EmpkinS-EKSpression randomized controlled trial (RCT; $N=256$, SCID-5-CV diagnoses) with the Extended Distress Analysis Interview Corpus (E-DAIC; $N=275$, semi-structured clinical interviews), predicting depression severity and binary diagnostic status from facial action units, head pose, and gaze. Cross-corpus binary classification proves more robust than continuous PHQ-8 severity regression, with forward transfer achieving AUC\,=\,0.70. Regression transfer is governed by functional context alignment: passive observation phases yield the most transferable models, while active emotion regulation phases elicit stronger within-corpus signals. These findings establish functional context alignment as the primary determinant of cross-corpus generalization, with passive
elicitation contexts offering the best trade-off between within-corpus
sensitivity and cross-corpus robustness.
\end{abstract}


\maketitle

\section{Introduction}

Depression is one of the most prevalent and disabling mental health conditions worldwide,
affecting more than 280 million people and accounting for a substantial share of
years lived with disability~\cite{who2023}. Despite its prevalence, depression
remains substantially under-diagnosed. Access to trained clinicians is limited,
stigma discourages help-seeking, and existing self-report instruments lack the
objectivity needed for population-scale monitoring. Automated analysis of
non-invasive behavioral signals, including speech acoustics and facial dynamics,
has emerged as a promising complementary approach for scalable, data-driven depression screening~\cite{cummins2015review,pampouchidou2017automatic, sadeghi2023exploring, sadeghi2024harnessing}. 

Facial action units (AUs), defined by the Facial Action Coding System~\cite{ekman1978}, provide a fine-grained, theory-grounded representation of facial muscle movement. In individuals with Major Depressive Disorder (MDD), psychomotor dysregulation manifests
as blunted expressivity, reduced AU activity, prolonged lip-corner depression (AU15, AU17), restricted gaze range, and decreased head-movement variability. These patterns have been associated with depression severity in both clinical and community samples~\cite{girard2014nonverbal, jain2014depression, SchullerRingeval2015}. Competitive challenges such as AVEC~\cite{ringeval2019avec} have pushed AU-based methods toward increasingly high within-corpus performance. Yet, translating these results to unseen datasets remains an open problem~\cite{cummins2015review, alghowinem2016cross}. 

A fundamental and often underappreciated source of that generalization gap is
\emph{elicitation context}. While studies have successfully identified facial
biomarkers of depression, such as reduced affiliative expressions and restricted
head motion, during naturalistic clinical interviews~\cite{girard2014nonverbal,
alghowinem2016multimodal, sadeghi2024harnessing}, these unstructured environments inherently lack the controlled, standardized emotional stimuli required to consistently elicit
pronounced psychomotor dysregulation. Crucially, structured and naturalistic paradigms are rarely compared within a single study, and with few recent exceptions exploring foundation models~\cite{dong2025foundation}, they are rarely tested in a cross-corpus transfer design~\cite{cummins2015review}. Consequently, a model trained on structured-task data may fail not because of domain shift in the feature distribution per se; instead, the underlying psychomotor signals are simply absent or attenuated in a naturalistic target corpus.

To address this context-dependency gap, we use the EmpkinS-\allowbreak EKSpression (EmpkinS) randomized controlled trial (RCT) dataset comprising 256 participants (128 MDD, 128 healthy controls) with rigorous Structured Clinical Interview for DSM-5 (SCID-5-CV) diagnostic ground truth, spanning four experimental phases and four intervention conditions. We characterize the within-corpus predictive value of facial features across 20 phase and condition combinations using four clinically validated severity scales: the Patient Health Questionnaire (PHQ-8 and PHQ-9), the Center for Epidemiological Studies Depression Scale (CES-D), and the 17-item Hamilton Rating Scale for Depression (HRSD-17). Severity is predicted most reliably during the passive Preparatory baseline phase, a counterintuitive result explained by the fact that active emotion regulation tasks introduce deliberate facial expression changes that mask the subtle psychomotor retardation characteristic of depression, while the passive phase leaves these trait-level differences undisturbed.

Building on this within-corpus analysis, we ask whether the most informative
phase and condition configurations generalize to an entirely independent corpus:
the Extended Distress Analysis Interview Corpus (E-DAIC)~\cite{gratch2014distress, devault2014simsensei}, collected under a semi-structured clinical
interview protocol with a different language and culture. We run systematic
bidirectional transfer experiments covering continuous PHQ-8 severity regression
and binary MDD classification, examining how transfer performance depends on
the functional alignment between elicitation contexts and the label instrument,
and whether passive observation phases yield more transferable models than
active emotion regulation phases.

These experiments yield four contributions. First, we conduct the first systematic
bidirectional cross-corpus study of facial biomarkers for depression, pairing
a structured clinical RCT with a naturalistic interview corpus for both
continuous severity regression and binary classification. Second, we provide
empirical evidence on the role of functional context alignment, label mismatch,
and feature space in determining cross-corpus generalization in facial depression
assessment. Third, we offer a controlled comparison of label settings
(SCID-5-CV vs.\ PHQ-8 threshold) to isolate the contribution of diagnostic protocol mismatch from contextual mismatch. Finally, we derive practical guidance for designing AI-based mental
health assessment protocols, highlighting the trade-offs between passive and
structured elicitation contexts for within-corpus sensitivity versus
cross-corpus robustness. To support full reproducibility, our code is publicly
available.\footnote{\url{https://github.com/MishaSadeghi/cross-corpus-facial-depression}}

\section{Related Work}
\label{sec:related}

AU-based facial features are well-established depression biomarkers, and the AVEC benchmark challenges~\cite{ringeval2019avec,valstar2016avec2016} have been pivotal in standardizing the field.
Key markers include AU15 (lip-corner depressor), AU17 (chin raiser), and reduced AU12 (lip-corner puller), all reliably associated with
MDD~\cite{girard2014nonverbal,jain2014depression,SchullerRingeval2015}. A combination of
AU intensities, gaze direction, and head pose predicts PHQ-8 or PHQ-9 scores
with CCC values of 0.3--0.6 within a single
corpus~\cite{valstar2016avec2016,ringeval2019avec,pampouchidou2017automatic}. Deep
learning has pushed within-corpus performance
higher~\cite{yang2017dcnn,zhu2017multimodal}, with recent work extending to
multimodal fusion with large language
models~\cite{sadeghi2024harnessing,sadeghi2023exploring}; however, greater model
capacity also heightens sensitivity to corpus-specific artefacts, making
cross-corpus generalization harder. Despite growing recognition of this problem, cross-corpus evaluations remain rare. \citet{cummins2015review} noted that the overwhelming majority of studies report exclusively within-corpus results, making it difficult to assess practical deployability. \citet{alghowinem2016cross} highlighted the difficulty of cross-cultural transfer, demonstrating that models relying on head pose and eye gaze require pooled multi-corpus training to learn universal nonverbal markers.
Speech-based cross-corpus transfer between a German
clinical corpus and E-DAIC has been demonstrated~\cite{sadeghi2026crosscorpus}, and recent foundation models have enabled lifespan-inclusive, multi-lingual cross-corpus evaluations for neuropsychiatric disorders~\cite{dong2025foundation}. To our knowledge, however, no prior work has isolated the cross-corpus generalization gap between a structured RCT paradigm and a naturalistic interview corpus using \emph{facial} features.

Facial affect expression is modulated by social display rules, cultural
background, and elicitation context~\cite{matsumoto2008cultural,
alghowinem2016multimodal}. AU-based depression markers are stronger in structured
paradigms than in free-conversation
settings~\cite{alghowinem2016multimodal,girard2014nonverbal}, and facial features reliably
track affective states such as stress during structured
tasks~\cite{rupp2025stress}. Cross-language transfer adds a further layer of
difficulty, as German and English-speaking populations may differ in baseline
expressivity norms, affecting the direction and magnitude of AU-based depression
indicators~\cite{matsumoto2008cultural}. Our work contributes controlled evidence
on all three axes: paradigm contrast (structured RCT vs.\ interview), language
(German vs.\ English), and label protocol (SCID-5-CV vs.\ PHQ-8 self-report). Label mismatch is a frequently cited but rarely quantified source of cross-corpus degradation~\cite{cummins2015review, alghowinem2016cross}. Studies vary in their depression target (PHQ-8, PHQ-9, BDI-II, HRSD, or clinician SCID diagnoses) and in binary classification thresholds~\cite{cummins2015review, alghowinem2016cross}. \citet{ringeval2019avec} noted that PHQ-8 and PHQ-9 are highly correlated, yet different binarization thresholds can
substantially alter class balance and apparent model performance. We address this
by introducing two label settings to disentangle label format mismatch from
contextual mismatch in our cross-corpus experiments.

\section{Methods} \label{sec:methods}

\begin{figure*}[t]
\centering
\includegraphics[width=1\textwidth]{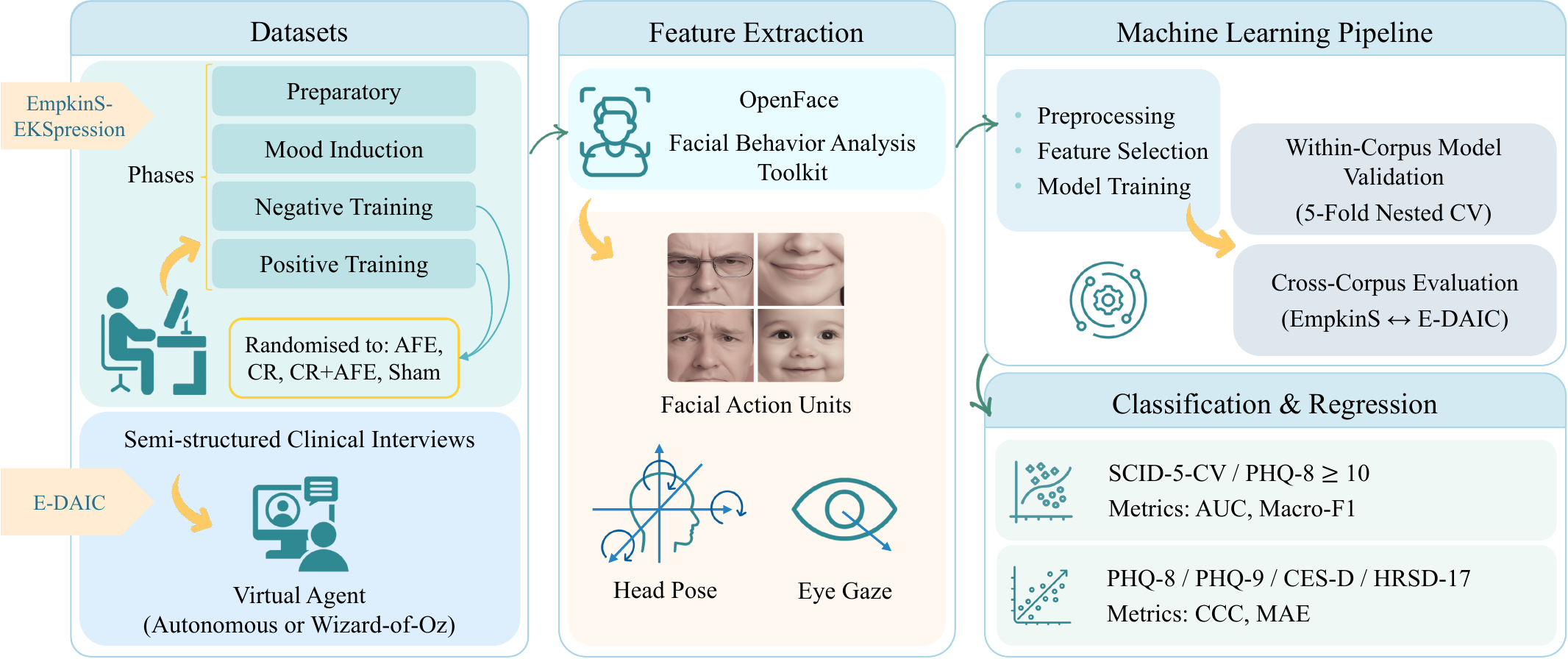}
\caption{Cross-corpus experimental pipeline pairing EmpkinS-EKSpression (EmpkinS) and
E-DAIC. OpenFace~\cite{baltrusaitis2018openface} features feed a unified framework evaluated within-corpus
and bidirectionally across corpora for severity regression and binary
classification.} 
\label{fig:pipeline}
\end{figure*}

Figure~\ref{fig:pipeline} illustrates the overall pipeline. Facial behavioral features from two corpora feed a unified machine learning framework evaluated within- and cross-corpus, targeting PHQ-8 severity regression and binary MDD classification. The two corpora are described below.

\subsection{Datasets}

\subsubsection{EmpkinS-\allowbreak EKSpression} 
This study leverages a German-lan\-guage corpus we collected during a four-\allowbreak arm, single-\allowbreak blind RCT~\cite{keinert2024facing} at the FAU Erlangen-\allowbreak Nürnberg EmpkinS Lab, centered on a smart\-phone-\allowbreak based training. The dataset comprises 256 participants, evenly split between 128 individuals with depressive disorders and 128 matched healthy controls. A defining feature of this dataset is its rigorous diagnostic ground truth: inclusion and binary classification labels are anchored in the SCID-5-CV, the clinical gold standard for psychiatric assessment. This distinguishes the corpus from many existing benchmarks on depression that rely solely on self-reported proxies. To enable continuous symptom severity estimation, participants underwent comprehensive assessments. These included self-reported measures such as the PHQ-8 (completed remotely prior to the lab visit), the PHQ-9, and the CES-D. Clinician-administered ratings were also collected, notably HRSD-17. Table \ref{tab:proposed_demographics} details the demographic and clinical characteristics of the cohort.

\begin{table}[ht]
\centering
\footnotesize
\caption{Demographics and clinical characteristics of the EmpkinS-EKSpression (EmpkinS) dataset. $p$-values: Mann--Whitney $U$ for continuous variables; $\chi^2$ for categorical.}
\label{tab:proposed_demographics}
\resizebox{\columnwidth}{!}{%
\begin{tabular}{lllr}
\toprule
\textbf{Characteristic} & \textbf{MDD$-$ ($n=128$)} & \textbf{MDD$+$ ($n=128$)} & $p$ \\
\midrule
\textit{Demographics} & & & \\
Age, mean ± SD & 36.7 ± 16.6 & 33.6 ± 13.4 & 0.622 \\
\quad 18–25 & 49 (38.3\%) & 44 (34.4\%) &  \\
\quad 26–35 & 27 (21.1\%) & 45 (35.2\%) &  \\
\quad $\geq$36 & 52 (40.6\%) & 39 (30.5\%) &  \\
Sex (F / M / Diverse) & 88 / 39 / 1 & 89 / 36 / 3 & 0.621 \\
RCT Condition$^\dagger$ & 32 / 32 / 32 / 32 & 32 / 32 / 32 / 32 & 1.000 \\
\midrule
\textit{Depression Severity Scores $^\ddagger$} & & & \\
\textit{PHQ-8} & & & \\
\quad Mean ± SD & 6.0 ± 4.1 & 14.1 ± 4.4 & < 0.001 \\
\quad Minimal (0–4) & 58 (45.3\%) & 2 (1.6\%) &  \\
\quad Mild (5–9) & 39 (30.5\%) & 18 (14.2\%) &  \\
\quad Moderate (10–14) & 26 (20.3\%) & 54 (42.5\%) &  \\
\quad Mod. Severe (15–19) & 5 (3.9\%) & 37 (29.1\%) &  \\
\quad Severe (20–24) & 0 (0.0\%) & 16 (12.6\%) &  \\
\textit{PHQ-9} & & & \\
\quad Mean ± SD & 4.3 ± 2.5 & 14.6 ± 4.8 & < 0.001 \\
\quad Minimal (0–4) & 70 (54.7\%) & 0 (0.0\%) &  \\
\quad Mild (5–9) & 57 (44.5\%) & 19 (15.0\%) &  \\
\quad Moderate (10–14) & 1 (0.8\%) & 45 (35.4\%) &  \\
\quad Mod. Severe (15–19) & 0 (0.0\%) & 45 (35.4\%) &  \\
\quad Severe (20–27) & 0 (0.0\%) & 18 (14.2\%) &  \\
\textit{CES-D} & & & \\
\quad Mean ± SD & 10.0 ± 6.8 & 28.0 ± 9.0 & < 0.001 \\
\quad Range & 0–32 & 8–49 &  \\
\textit{HRSD-17} & & & \\
\quad Mean ± SD & 4.8 ± 4.0 & 16.1 ± 6.2 & < 0.001 \\
\bottomrule
\multicolumn{4}{l}{%
  \parbox{\linewidth}{\vspace{4pt}\footnotesize
    $^\dagger$ RCT: randomized controlled trial. Conditions: AFE = Anti-Depressive
    Facial Expression; CR = Cognitive Reappraisal; CR+AFE = CR and AFE Combined;
    SHAM~=~Sham~Control.\\[2pt]
    $^\ddagger$ PHQ-8/9: Patient Health Questionnaire (8/9); CES-D: Center for
    Epidemiological Studies Depression Scale; HRSD-17: Hamilton Rating Scale for
    Depression \mbox{(17-item)}.
  }%
}\\
\end{tabular}%
}
\end{table}

\paragraph{Intervention Conditions and Experimental Phases}
Participants completed a laboratory session guided by a custom smartphone application (EmpkinS-EKSpression app). The session was structured into specific experimental phases designed to systematically capture facial dynamics across both passive observation and active emotion regulation tasks:

\begin{itemize}
    \item Preparatory Phase: A baseline period to capture initial affective states. Participants were passively exposed to alternating self-recorded videos and emotionally valenced words (such as ``hopeless'' or ``happy'', drawn from Beck's cognitive triad~\cite{beck1979cognitive}) on the smartphone screen, providing a resting state comparison.
    \item Mood Induction: To establish a uniform affective baseline before the active interventions, participants completed a guided task to induce a transient depressive mood~\cite{keinert2024facing}. They read negatively valenced, self-referential cognitions aloud while listening to melancholic music.
    \item Training Phases: The core of the intervention consisted of two active blocks of 20 randomized trials each, specifically divided into a Negative Training phase (targeting depressogenic content) and a Positive Training phase (reinforcing anti-depressive content). During these phases, participants actively practiced specific cognitive or expressive emotion regulation strategies derived from validated psychological instruments~\cite{kliem2018bhs, possel2005automatische, ferring1996rosenberg, keinert2024facing}.  
\end{itemize}

During the training phases, participants were randomized into one of four conditions to isolate different mechanisms of affect regulation:

\begin{itemize}
    \item Cognitive Reappraisal (CR): A purely cognitive task where participants engaged in structured verbal restructuring. During Negative Training, they completed a three-step verbal response process to explicitly reject the negative thought, affirm a positive alternative, and state a self-sup\-port\-ing phrase. During Positive Training, they applied a congruent verbal strategy to actively endorse and internalize anti-depressive statements.    
    \item Anti-Depressive Facial Expression (AFE): An isolated motor-expressive task without verbal reappraisal. Participants were required to actively modify their facial expressions to either counteract depressogenic statements or reinforce anti-depressive ones. This targeted specific facial muscles (e.g., corrugator supercilii and zygomaticus major) and utilized real-time visual biofeedback adjusted by a remote therapist.   
    \item Combined Training (CR+AFE): An integrated condition that merged the verbal cognitive restructuring of the CR group with congruent bodily and facial expressions. Participants aligned their verbal reappraisal statements with specific physical postures, including rejecting gestures for negative cognitions, and positive affective facial expressions for positive ones.    
    \item Sham Control: A neutral comparative condition consisting of geometric object matching tasks. This controlled for the nonspecific effects of smartphone interaction, visual feedback, and general psychomotor engagement.
\end{itemize}

Throughout all experimental phases within the controlled laboratory setting, multimodal behavioral and psychophysiological data were continuously captured. For comprehensive details regarding the experimental setup and the full scope of collected modalities, readers are referred to the study protocol~\cite{keinert2024facing}. The present analysis focuses entirely on visual behavioral biomarkers extracted from the frontal smartphone video recordings.

\subsubsection{E-DAIC Corpus}
As the target for our cross-corpus evaluation, we utilize the E-DAIC~\cite{gratch2014distress, devault2014simsensei} dataset. In contrast to the structured emotion regulation tasks of the EmpkinS dataset, E-DAIC provides a conversational, semi-clinical interview setting. These sessions, lasting approximately 20 minutes, are conducted by a virtual agent (Ellie) operating either autonomously or via a Wizard-of-Oz setup. The dataset comprises 275 interview sessions partitioned into official training ($n=163$), development ($n=56$), and test ($n=56$) subsets. Table \ref{tab:edaic_demographics} outlines the demographic distribution and dataset partitions. Depression severity is quantified using self-reported PHQ-8 scores (ranging from 0 to 24), and a threshold of PHQ-8\,$\geq 10$ is employed to designate the depressed class for binary classification tasks. E-DAIC exhibits a class imbalance, with a roughly 3.2:1 ratio of healthy to depressed participants, with the majority of instances clustering at the lower end of the PHQ-8 scale. This right-skewed distribution, where severe depression cases are rare, introduces inherent challenges for training robust continuous severity regression models. The dataset is distributed with raw audio recordings and transcripts, alongside pre-extracted visual behavioral features including facial action units, head pose, and eye gaze.

\begin{table}[ht]
\centering
\footnotesize
\caption{Demographic and clinical characteristics of the \mbox{E-DAIC} dataset. Depression is defined as a PHQ-8 $\geq 10$. $p$-values: Mann--Whitney $U$ for continuous variables; $\chi^2$ for categorical.}
\label{tab:edaic_demographics}
\begin{tabular}{lllr}
\toprule
\textbf{Characteristic} & \textbf{MDD$-$ ($n=209$)} & \textbf{MDD$+$ ($n=66$)} & $p$ \\
\midrule
\textit{Demographics} & & & \\
Age, mean ± SD & 41.1 ± 13.0 & 41.6 ± 11.8 & 0.697 \\
\quad 18–25 & 29 (13.9\%) & 6 (9.1\%) &  \\
\quad 26–35 & 55 (26.3\%) & 19 (28.8\%) &  \\
\quad $\geq$36 & 125 (59.8\%) & 41 (62.1\%) &  \\
Sex (F / M / Unknown) & 74 / 134 / 1 & 31 / 35 / 0 & 0.130 \\
\textit{Official Dataset Split} & & & \\
\quad Train & 126 (60.3\%) & 37 (56.1\%) &  \\
\quad Development & 44 (21.1\%) & 12 (18.2\%) &  \\
\quad Test & 39 (18.7\%) & 17 (25.8\%) &  \\
\midrule
\textit{Depression Severity (PHQ-8)$^\dagger$} & & & \\
Score, mean ± SD & 4.2 ± 3.7 & 15.7 ± 3.5 & < 0.001 \\
\quad Minimal (0–4) & 122 (58.4\%) & 0 (0.0\%) &  \\
\quad Mild (5–9) & 67 (32.1\%) & 0 (0.0\%) &  \\
\quad Moderate (10–14) & 18 (8.6\%) & 25 (37.9\%) &  \\
\quad Mod. Severe (15–19) & 2 (1.0\%) & 31 (47.0\%) &  \\
\quad Severe (20–24) & 0 (0.0\%) & 10 (15.2\%) &  \\
\bottomrule 
\multicolumn{4}{l}{%
  \parbox{\linewidth}{\vspace{4pt}\footnotesize
    $^\dagger$ PHQ-8: Patient Health Questionnaire-8.
  }%
}
\end{tabular}%
\end{table}

\subsection{Feature Extraction and Alignment}
\label{sec:features}

The EmpkinS corpus was processed with OpenFace~2.2.0~\cite{baltrusaitis2018openface, baltruvsaitis2016openface},
yielding per-frame estimates of AU intensities and presence indicators, bilateral gaze direction vectors, gaze angles, and head pose (translation and
rotation). Low-quality frames (OpenFace confidence\,$< 0.5$) were discarded. Five participants were excluded due to consent withdrawal, technical
recording failures, or insufficient retained frames ($< 50$),
yielding 251 participants for analysis.
The E-DAIC corpus provides equivalent features as pre-extracted outputs distributed with the dataset, computed using the OpenFace toolkit~\cite{baltruvsaitis2016openface} by the corpus curators. Each signal column was summarized into a fixed-length feature vector using 18 statistical functionals applied to both the original signal and its first-order frame differences: mean, std, min, max, skew, kurtosis, range, and Shannon entropy for each; plus rate-of-change and peak count on the original signal. After applying this aggregation to both corpora, we intersected their column sets to obtain 882 shared features covering all AUs, gaze angles, and head pose rotations, while excluding corpus-specific landmark positions and composite expressions absent from E-DAIC.

\begin{table*}
\centering
\small
\setlength{\tabcolsep}{4pt}    
\renewcommand{\arraystretch}{1.1}
\caption{Within-corpus regression results. Cells report the mean $\pm$ standard
deviation across folds for the model achieving the highest mean CCC. Bold = best CCC within each condition across phases. Underline = best overall CCC per scale.}
\label{tab:within}
\begin{tabular}{l rr rr rr rr rr}
\toprule
& \multicolumn{2}{c}{\textbf{AFE}}
  & \multicolumn{2}{c}{\textbf{CR+AFE}}
  & \multicolumn{2}{c}{\textbf{CR}}
  & \multicolumn{2}{c}{\textbf{SHAM}} 
  & \multicolumn{2}{c}{\textbf{All Conditions}} \\
\cmidrule(lr){2-3}\cmidrule(lr){4-5}\cmidrule(lr){6-7}\cmidrule(lr){8-9}\cmidrule(lr){10-11}
\textbf{Target / Phase} & CCC & MAE & CCC & MAE & CCC & MAE & CCC & MAE & CCC & MAE \\
\midrule    
\multicolumn{11}{l}{\textbf{PHQ-8}} \\
\quad Preparatory   & \underline{\textbf{0.31$\pm$0.22}} & 5.34$\pm$1.51 & \textbf{0.17$\pm$0.11} & 5.03$\pm$0.60 & 0.00$\pm$0.20 & 5.83$\pm$1.19 & \textbf{0.28$\pm$0.17} & 4.85$\pm$0.98  & 0.04$\pm$0.15 & 5.11$\pm$0.65 \\
\quad Mood Induction       & $-$0.01$\pm$0.08 & 7.34$\pm$2.45 & 0.02$\pm$0.15 & 5.54$\pm$1.30 & 0.09$\pm$0.24 & 5.18$\pm$1.17 & 0.03$\pm$0.12 & 6.05$\pm$1.50  & \textbf{0.14$\pm$0.05} & 5.10$\pm$0.27 \\
\quad Negative Training    & 0.19$\pm$0.18 & 5.14$\pm$0.68 & 0.09$\pm$0.26 & 4.86$\pm$1.25 & \textbf{0.20$\pm$0.20} & 5.11$\pm$0.99 & 0.26$\pm$0.16 & 5.03$\pm$0.96 & 0.13$\pm$0.10 & 5.26$\pm$0.60 \\
\quad Positive Training     & 0.26$\pm$0.33 & 5.19$\pm$1.48 & $-$0.04$\pm$0.14 & 5.93$\pm$0.96 & 0.18$\pm$0.24 & 5.90$\pm$1.29 & 0.24$\pm$0.23 & 4.87$\pm$1.05 & 0.09$\pm$0.05 & 5.08$\pm$0.49 \\
\midrule

\multicolumn{11}{l}{\textbf{PHQ-9}} \\
\quad Preparatory           & \underline{\textbf{0.34$\pm$0.14}} & 5.33$\pm$1.95 & 0.03$\pm$0.11 & 8.17$\pm$3.92 & \textbf{0.34$\pm$0.18} & 5.24$\pm$0.85 & 0.22$\pm$0.26 & 5.14$\pm$1.50 & 0.12$\pm$0.07 & 5.65$\pm$0.20 \\
\quad Mood Induction       & $-$0.01$\pm$0.30 & 6.02$\pm$1.02 & 0.13$\pm$0.10 & 7.02$\pm$1.46 & 0.00$\pm$0.24 & 6.76$\pm$1.16 & 0.17$\pm$0.13 & 5.83$\pm$1.04 & 0.12$\pm$0.08 & 5.47$\pm$0.43 \\
\quad Negative Training      & 0.20$\pm$0.18 & 5.31$\pm$0.80 & 0.07$\pm$0.24 & 6.06$\pm$1.41 & 0.20$\pm$0.25 & 5.71$\pm$1.02 & 0.28$\pm$0.11 & 5.00$\pm$0.99 & 0.11$\pm$0.11 & 5.55$\pm$0.50 \\
\quad Positive Training       & $-$0.11$\pm$0.19 & 6.40$\pm$1.05 & \textbf{0.17$\pm$0.15} & 6.56$\pm$1.63 & 0.21$\pm$0.25 & 5.88$\pm$1.18 & \textbf{0.33$\pm$0.08} & 5.00$\pm$0.79 & \textbf{0.14$\pm$0.15} & 5.36$\pm$0.58 \\
\midrule

\multicolumn{11}{l}{\textbf{CES-D}} \\
\quad Preparatory           & \textbf{0.17$\pm$0.35} & 10.61$\pm$2.46 & 0.03$\pm$0.17 & 11.78$\pm$2.17 & 0.07$\pm$0.26 & 11.91$\pm$1.49 & 0.35$\pm$0.12 & 9.09$\pm$2.13 & \textbf{0.14$\pm$0.06} & 10.03$\pm$0.43\\
\quad Mood Induction       & 0.09$\pm$0.10 & 11.93$\pm$1.40 & \textbf{0.10$\pm$0.30} & 10.35$\pm$3.08 & 0.16$\pm$0.16 & 12.22$\pm$1.82 & 0.03$\pm$0.29 & 11.51$\pm$2.16 & 0.11$\pm$0.08 & 10.45$\pm$1.48 \\
\quad Negative Training      & 0.13$\pm$0.17 & 9.94$\pm$2.11 & $-$0.03$\pm$0.27 & 12.43$\pm$1.52 & 0.18$\pm$0.16 & 10.58$\pm$1.92 & 0.20$\pm$0.15 & 10.15$\pm$0.78 & 0.07$\pm$0.08 & 10.78$\pm$0.90 \\
\quad Positive Training       & $-$0.08$\pm$0.24 & 12.33$\pm$2.12 & 0.07$\pm$0.17 & 10.57$\pm$2.73 & \textbf{0.20$\pm$0.32} & 10.59$\pm$1.97 & \underline{\textbf{0.36$\pm$0.23}} & 9.26$\pm$2.29 & 0.05$\pm$0.13 & 10.55$\pm$1.45 \\
\midrule

\multicolumn{11}{l}{\textbf{HRSD-17}} \\
\quad Preparatory           & 0.06$\pm$0.24 & 7.29$\pm$1.95 & $-$0.10$\pm$0.32 & 6.99$\pm$1.49 & \textbf{0.22$\pm$0.11} & 7.61$\pm$0.71 & 0.18$\pm$0.19 & 6.34$\pm$1.62 & \textbf{0.10$\pm$0.02} & 6.66$\pm$0.55\\
\quad Mood Induction       & 0.01$\pm$0.20 & 8.96$\pm$1.66 & 0.24$\pm$0.29 & 6.84$\pm$2.62 & 0.12$\pm$0.24 & 7.61$\pm$2.26 & 0.10$\pm$0.12 & 6.23$\pm$1.14 & 0.05$\pm$0.16 & 6.96$\pm$1.18\\
\quad Negative Training      & $-$0.03$\pm$0.20 & 8.53$\pm$1.43 & 0.18$\pm$0.19 & 6.38$\pm$1.60 & 0.08$\pm$0.20 & 7.88$\pm$0.88 & 0.25$\pm$0.12 & 6.39$\pm$0.98 & 0.07$\pm$0.12 & 6.69$\pm$0.24\\
\quad Positive Training       & \textbf{0.12$\pm$0.15} & 7.16$\pm$1.85 & \underline{\textbf{0.29$\pm$0.13}} & 6.36$\pm$1.46 & 0.19$\pm$0.27 & 7.72$\pm$2.16 & \textbf{0.26$\pm$0.23} & 5.71$\pm$1.59 & 0.10$\pm$0.10 & 6.42$\pm$0.25\\
\bottomrule
\end{tabular}
\end{table*}

\subsection{Machine Learning Pipeline}
\label{sec:pipeline}

For EmpkinS we use pre-computed stratified 80/20 splits (stratified
by diagnosis and condition), shared across all phase and condition combinations
to ensure comparability in cross-corpus settings. For E-DAIC we use the official
train+dev\,/\,test partition. All preprocessing steps are fit exclusively on the
training partition and applied to the test set without refitting. Missing values,
arising from participants excluded in a given phase, are imputed with the
training-set column median.

Feature selection proceeds in two steps, both computed on the training set only.
First, features are filtered by univariate association with the target
(Spearman $|\rho|$ for regression, Mann--Whitney $U$ for classification),
retaining those surviving Benjamini--Hochberg FDR correction ($q = 0.20$),
with a fallback of the top 20 when fewer than 20 pass the threshold. Second,
recursive feature elimination (RFE) with a Ridge surrogate model further removes
redundant features (details in Appendix~\ref{app:hyperparams}).

For regression we evaluate 12 models: Ridge, Lasso, ElasticNet, Support Vector Regression (SVR), K-Nearest Neighbors ($k$-NN), Random Forest, Extra Trees, Gradient Boosting, XGBoost~\cite{chen2016xgboost}, LightGBM~\cite{ke2017lightgbm}, CatBoost~\cite{prokhorenkova2018catboost}, and Multilayer Perceptron (MLP).
For classification we evaluate Logistic Regression, Support Vector Classifier~(SVC), Random Forest, Extra Trees, Gradient Boosting, XGBoost, LightGBM, CatBoost, and MLP. 
Three scalers (Standard, MinMax, Robust) are evaluated
jointly with each model. Hyperparameters are selected via 5-fold grid search on the
training set (full grids in Appendix~\ref{app:hyperparams}).

We report CCC (Lin's concordance correlation coefficient~\cite{lin1989ccc}) as the
primary regression metric, as it jointly penalizes poor rank-order correlation and
systematic scale or offset bias, a property particularly relevant for cross-corpus
evaluation where the latter is common. Pearson $r$, MAE, and RMSE are reported as
secondary metrics. For classification, AUC is the primary metric; macro-F1 is reported
as secondary, given the class imbalance in E-DAIC.

\subsection{Experimental Designs}
\label{sec:experiments}

\subsubsection{Within-Corpus Regression}
\label{sec:within}

To identify which phase and condition combinations are most informative for depression severity estimation, we apply 5-fold nested cross-validation across all 4 phases $\times$ 5 conditions (AFE, CR, CR+AFE, SHAM, and all conditions pooled), yielding 20 combinations in total. We evaluate four clinically relevant scales (PHQ-8, PHQ-9, CES-D, HRSD-17) to characterize target reliability.

\subsubsection{Cross-Corpus Regression (EmpkinS $\leftrightarrow$ E-DAIC)}
\label{sec:cross}

We design two directional transfer experiments using the same aligned 882-feature
set. In the forward direction, models are trained on EmpkinS 80\% training split and tested on the E-DAIC official test set, assessing transfer from the RCT to the clinical interview. In the reverse direction, models are
trained on E-DAIC train+dev and tested on EmpkinS 20\% holdout, examining whether interview-trained models generalize to RCT phases. In both directions, preprocessing and feature selection are fit on the source partition only and applied to the target without refitting. Within-corpus baselines (EmpkinS\,$\to$\,EmpkinS and E-DAIC\,$\to$\,E-DAIC) are trained and tested under the same fixed splits to anchor performance.

\subsubsection{Cross-Corpus Classification (EmpkinS $\leftrightarrow$ E-DAIC)}
\label{sec:cross_cls}

The same bidirectional design and within-corpus baselines are applied to binary depression classification (individuals with MDD vs.\ healthy controls). We additionally train on \emph{all} EmpkinS participants and test on the E-DAIC official test set, exploiting the maximum available RCT training data when no within-corpus test set is required. Furthermore, we evaluate two label settings to isolate the effect of label mismatch. \emph{Setting~1} uses SCID-5-CV binary labels for EmpkinS and PHQ-8\,$\geq 10$ for E-DAIC (diagnostic protocol mismatch). \emph{Setting~2} uses PHQ-8\,$\geq 10$ for both corpora (matched label definition). 

\begin{table*}[t]
\centering
\small
\setlength{\tabcolsep}{6pt}  
\caption{Cross-corpus PHQ-8 regression. Bold: best CCC within each experiment row. Underline: best CCC within each transfer direction.}
\label{tab:cc_reg}
\begin{tabular}{l *{14}{r}}
\toprule
& \multicolumn{6}{c}{\textbf{Preparatory phase}}
& \multicolumn{8}{c}{\textbf{Best within-corpus phase per condition}} \\
\cmidrule(lr){2-7}\cmidrule(lr){8-15}
& \multicolumn{2}{c}{\textbf{Prep./AFE}}
& \multicolumn{2}{c}{\textbf{Prep./SHAM}}
& \multicolumn{2}{c}{\textbf{Prep./All}}
& \multicolumn{2}{c}{\textbf{Pos./AFE}}
& \multicolumn{2}{c}{\textbf{Neg./CR+AFE}}
& \multicolumn{2}{c}{\textbf{Neg./CR}}
& \multicolumn{2}{c}{\textbf{Neg./SHAM}} \\
\cmidrule(lr){2-3}\cmidrule(lr){4-5}\cmidrule(lr){6-7}
\cmidrule(lr){8-9}\cmidrule(lr){10-11}\cmidrule(lr){12-13}\cmidrule(lr){14-15}
& CCC & MAE & CCC & MAE & CCC & MAE
& CCC & MAE & CCC & MAE & CCC & MAE & CCC & MAE \\
\midrule
\multicolumn{15}{c}{\textbf{Cross-corpus}} \\
\midrule
EmpkinS$^\dagger_{\rm all}$ $\to$ E-DAIC
  & 0.03 & 6.13 & 0.00 & 5.83 & 0.00 & 7.18
  & $-$0.03 & 6.86 & 0.13 & 5.62 & \textbf{0.15} & 5.77 & 0.10 & 5.49 \\
EmpkinS $\to$ E-DAIC
  & 0.01 & 6.98 & 0.08 & 5.40 & 0.03 & 9.59
  & 0.04 & 8.76 & 0.03 & 8.75 & \underline{\textbf{0.21}} & 8.87 & 0.03 & 6.27 \\
E-DAIC $\to$ EmpkinS
  & 0.14 & 6.83 & \underline{\textbf{0.26}} & 5.50 & 0.10 & 4.77
  & 0.10 & 7.25 & 0.25 & 5.40 & 0.04 & 5.31 & 0.14 & 5.06 \\
\midrule
\multicolumn{15}{c}{\textbf{Within-corpus baseline}} \\
\midrule
EmpkinS $\to$ EmpkinS
  & 0.03 & 6.42 & 0.19 & 6.40 & 0.09 & 4.63
  & 0.10 & 6.00 & 0.05 & 6.25 & \textbf{0.28} & 3.91 & 0.14 & 4.78 \\
E-DAIC $\to$ E-DAIC$\,\ddagger$
  & 0.08 & 5.39 & 0.08 & 5.39 & 0.08 & 5.39
  & 0.08 & 5.39 & 0.08 & 5.39 & 0.08 & 5.39 & 0.08 & 5.39 \\
\bottomrule
\multicolumn{15}{l}{\footnotesize $^\dagger$ EmpkinS = EmpkinS-EKSpression. $^\ddagger$ E-DAIC within-corpus result is config-independent.}
\end{tabular}
\end{table*}

\section{Results and Discussion}
\label{sec:results}

\subsection{Within-Corpus Regression}
\label{sec:results_within}

Table~\ref{tab:within} summarizes the 5-fold nested cross-validation performance across all 20 phase and condition combinations and four depression scales. We consider the PHQ-8 as our primary regression target to enable direct cross-corpus evaluations with the E-DAIC dataset, which only provides PHQ-8 scores. However, it is important to note that the other in-lab assessments, such as the CES-D and HRSD-17, offer higher clinical validity compared to the remotely collected PHQ-8~\cite{cuijpers2010self}, as clinician-administered instruments have been shown to yield more reliable severity estimates than self-report
proxies, particularly when the latter are completed outside a supervised clinical setting. 
For the PHQ-8, the CCC peaks at 0.31 in the Preparatory/AFE setting and 0.28 in Preparatory/SHAM. This suggests that for this specific self-report scale, the passive baseline phase carries a highly reliable signal. During the Preparatory phase, minimal behavioral demands are placed on participants. This lack of intervention may allow trait-level psychomotor differences between the MDD and healthy control groups to surface without the confounding effects of structured tasks. However, this passive superiority is not a universal rule across all scales. During the temporal blocks designated for active training (Negative and Positive Training), specific subgroups yield comparable or superior predictive power. Most notably, the SHAM condition frequently achieves the highest performance during these phases, reaching a CCC of 0.36 for the CES-D and 0.33 for the PHQ-9. SHAM is a neutral, active control condition involving geometric object matching. This finding reveals that imposing a standardized cognitive load, without explicitly instructing participants to alter their emotional or facial expressions, may provide an optimal environment for extracting depressive biomarkers. Active regulation tasks, such as AFE, require participants to artificially manipulate their facial muscles, which might inadvertently mask subtle, trait-level psychomotor retardation or blunted affect. 

Furthermore, examining the aggregated ``All Conditions'' columns reveals that pooling data across all four interventions often diminishes performance. For example, the PHQ-8 Preparatory phase drops to a CCC of 0.04 when all conditions are combined, despite this setting providing the largest training set ($n=251$). This pattern recurs across various phases and underscores that merging disparate interaction contexts actively dilutes condition-specific facial signals. Condition homogeneity appears more critical than sample size for within-corpus continuous regression.

While the self-report measures (PHQ-8, PHQ-9, and CES-D) occasionally demonstrate overlapping areas of predictive success, such as within the SHAM condition and the Negative Training phase, their performance trajectories do not perfectly align. Notable variances exist across conditions and phases, indicating that isolated facial features interact differently with the specific symptom weightings of each questionnaire. Conversely, the HRSD-17 shows a different correlation pattern altogether. Since the HRSD-17 relies on an expert rater synthesizing holistic behavioral cues alongside specific physical symptoms, it inherently diverges from patterns driven purely by self-perception and isolated facial dynamics.

Overall, the absolute CCC values across all target variables remain modest. This outcome reflects the inherent difficulty of continuously predicting subjective depression severity from isolated facial features, alongside the strict evaluation criteria of nested cross-validation, which naturally yields more conservative estimates than simple holdout techniques. Establishing these performance limits provides a realistic baseline required to accurately interpret the cross-corpus transfer results in the subsequent sections.

\begin{table*}[t]
\centering
\small
\setlength{\tabcolsep}{9pt}  
\caption{Cross-corpus binary depression classification. AUC and macro-averaged F1 per configuration and label setting. \\
Bold: best AUC within each experiment row. Underline: best AUC within each transfer direction.}
\label{tab:cc_cls}
\begin{tabular}{l c@{\hspace{5pt}}c c@{\hspace{5pt}}c c@{\hspace{5pt}}c c@{\hspace{5pt}}c c@{\hspace{5pt}}c c@{\hspace{5pt}}c c@{\hspace{5pt}}c}
\toprule
& \multicolumn{6}{c}{\textbf{Preparatory phase}} & \multicolumn{8}{c}{\textbf{Best within-corpus phase per condition}} \\
\cmidrule(lr){2-7}\cmidrule(lr){8-15}
& \multicolumn{2}{c}{\textbf{Prep./AFE}} & \multicolumn{2}{c}{\textbf{Prep./SHAM}}
& \multicolumn{2}{c}{\textbf{Prep./All}} & \multicolumn{2}{c}{\textbf{Pos./AFE}}
& \multicolumn{2}{c}{\textbf{Neg./CR+AFE}} & \multicolumn{2}{c}{\textbf{Neg./CR}}
& \multicolumn{2}{c}{\textbf{Neg./SHAM}} \\
\cmidrule(lr){2-3}\cmidrule(lr){4-5}\cmidrule(lr){6-7}
\cmidrule(lr){8-9}\cmidrule(lr){10-11}\cmidrule(lr){12-13}\cmidrule(lr){14-15}
& AUC & F1$_{\rm M}$ & AUC & F1$_{\rm M}$ & AUC & F1$_{\rm M}$ & AUC & F1$_{\rm M}$ & AUC & F1$_{\rm M}$ & AUC & F1$_{\rm M}$ & AUC & F1$_{\rm M}$ \\
\midrule
\multicolumn{15}{c}{\textbf{Setting 1 -- EmpkinS$^\dagger$: SCID labels vs. E-DAIC: PHQ-8$\,\geq 10$ labels}} \\
\midrule
EmpkinS$_{\rm all}$ $\to$ E-DAIC
  & 0.62 & 0.30 & 0.60 & 0.50 & \underline{\textbf{0.70}} & 0.59
  & 0.55 & 0.28 & 0.67 & 0.42 & 0.57 & 0.31 & 0.66 & 0.41 \\
EmpkinS $\to$ E-DAIC
  & 0.59 & 0.50 & 0.55 & 0.41 & 0.57 & 0.23
  & 0.58 & 0.58 & 0.57 & 0.48 & \textbf{0.65} & 0.46 & 0.59 & 0.41 \\
E-DAIC $\to$ EmpkinS
  & 0.88 & 0.33 & 0.79 & 0.58 & 0.57 & 0.52
  & \underline{\textbf{1.00}} & 0.33 & 0.67 & 0.50 & 0.69 & 0.44 & 0.74 & 0.58 \\
EmpkinS $\to$ EmpkinS
  & 0.67 & 0.33 & 0.36 & 0.31 & 0.66 & 0.67
  & 0.47 & 0.50 & 0.47 & 0.56 & \textbf{0.78} & 0.62 & 0.64 & 0.45 \\
E-DAIC $\to$ E-DAIC$\,\ddagger$
  & 0.58 & 0.50 & 0.58 & 0.50 & 0.58 & 0.50
  & 0.58 & 0.50 & 0.58 & 0.50 & 0.58 & 0.50 & 0.58 & 0.50 \\
\midrule
\multicolumn{15}{c}{\textbf{Setting 2 -- PHQ-8$\,\geq 10$ labels for both corpora}} \\
\midrule
EmpkinS$_{\rm all}$ $\to$ E-DAIC
  & 0.57 & 0.23 & \underline{\textbf{0.69}} & 0.35 & 0.66 & 0.23
  & 0.62 & 0.23 & 0.61 & 0.59 & 0.64 & 0.29 & 0.67 & 0.29 \\
EmpkinS $\to$ E-DAIC
  & 0.61 & 0.39 & 0.53 & 0.50 & \textbf{0.63} & 0.30
  & 0.59 & 0.37 & 0.58 & 0.45 & 0.57 & 0.23 & 0.54 & 0.46 \\
E-DAIC $\to$ EmpkinS
  & 0.78 & 0.50 & 0.67 & 0.19 & 0.57 & 0.30
  & 0.78 & 0.25 & 0.77 & 0.69 & 0.63 & 0.49 & \underline{\textbf{0.85}} & 0.68 \\
EmpkinS $\to$ EmpkinS
  & 0.48 & 0.50 & 0.83 & 0.46 & 0.60 & 0.65
  & 0.63 & 0.56 & 0.60 & 0.31 & \textbf{0.91}& 0.70 & 0.57 & 0.41 \\
E-DAIC $\to$ E-DAIC$\,\ddagger$
  & 0.58 & 0.50 & 0.58 & 0.50 & 0.58 & 0.50
  & 0.58 & 0.50 & 0.58 & 0.50 & 0.58 & 0.50 & 0.58 & 0.50 \\
\bottomrule
\multicolumn{15}{l}{\footnotesize $^\dagger$ EmpkinS = EmpkinS-EKSpression. $^\ddagger$ E-DAIC within-corpus result is config-independent.}
\end{tabular}
\end{table*}

\subsection{Cross-Corpus Regression}
\label{sec:results_cross}

To ensure a focused cross-corpus evaluation, Table~\ref{tab:cc_reg} restricts reporting to seven specific source configurations. These include three variants of the Preparatory phase: AFE and SHAM, selected for their strong passive baselines, alongside a pooled Preparatory/All condition designed to maximize available training data. Furthermore, we include four active phase and condition combinations (Pos./AFE, Neg./CR+AFE, Neg./CR, and Neg./SHAM) previously identified as the most discriminative during the within-corpus analysis.

Evaluating the transferability between these datasets reveals varying degrees of generalization depending on the transfer direction. When transferring models trained on the semi-structured clinical interviews to the structured tasks (E-DAIC $\to$ EmpkinS), the predictive performance demonstrates notable success. Specifically, the CCC reaches 0.26 for the Prep./SHAM configuration and 0.25 for Neg./CR+AFE. These values closely approach the optimal within-corpus EmpkinS baseline, which peaks at 0.28 for Neg./CR. The success in the Preparatory phase aligns with expectations, as passive observation shares behavioral similarities with the naturalistic listening segments of a semi-structured interview. Simultaneously, the transferability to Neg./CR+AFE suggests that the active elicitation of negative emotions combined with expressive regulation in the RCT may mirror moments of intense emotional disclosure during the E-DAIC interview. Conversely, the pooled Prep./All condition drops to 0.10, consistent with the condition-dilution effect observed during the within-corpus analysis.

In the forward transfer direction, predicting E-DAIC scores using models trained on the EmpkinS dataset presents greater challenges. Training on the entire EmpkinS cohort (EmpkinS$_{\rm all}$ $\to$ E-DAIC) yields a maximum CCC of 0.15 for the Neg./CR configuration. Restricting the source to the 80\% training split (EmpkinS $\to$ E-DAIC) produces a similar peak of 0.21 for Neg./CR, while other configurations hover near zero. Since utilizing all EmpkinS participants does not improve upon the split-based evaluation, this attenuation is likely not attributable to insufficient training data. Instead, the primary driver appears to be contextual mismatch. The facial dynamics elicited during structured affective tasks, such as erratic AU intensity peaks or restricted gaze entropy, represent specific signatures of psychomotor dysregulation under affective challenge. These context-specific dynamics do not naturally manifest during a clinical interview. Consequently, a model trained heavily on challenge-induced patterns finds limited corresponding signals in the E-DAIC footage.

The modest CCC observed in the Neg./CR configuration (0.21) during split-based forward transfer may reflect a coincidental functional alignment. The negative cognitive restructuring task involves sustained verbal processing of self-referential negative content. This shares characteristics with the conversational demands of the E-DAIC interview, where participants frequently discuss depressive symptoms with a virtual agent. Both contexts require structured verbal engagement that may elicit partially overlapping facial inhibition patterns in depressed individuals. However, without further feature-level analysis, this interpretation remains speculative. Finally, the E-DAIC $\to$ E-DAIC within-corpus baseline remains notably low at 0.08. This reflects the inherent difficulty of continuous PHQ-8 regression from isolated facial features in a semi-structured conversational setting, where facial movements driven by speech often obscure subtle affective signals. Furthermore, this difficulty is severely compounded by the right-skewed severity distribution. Since the vast majority of E-DAIC participants cluster at the lower end of the scoring scale, regression models tend to predict the mean severity to minimize absolute error. This lack of predictive variance heavily penalizes the CCC, leading to the near-chance CCC baseline. 

\subsection{Cross-Corpus Classification}
\label{sec:results_cls}

To ensure a consistent evaluation framework, the same seven source configurations used in the regression analysis are utilized in the classification experiments. Table~\ref{tab:cc_cls} reports the AUC and macro-averaged F1 (F1$_{\rm M}$) scores for binary depression classification across two label settings. In contrast to the near-complete predictive collapse observed during continuous regression, categorical classification demonstrates notable cross-corpus robustness.

In Setting 1 (incorporating a diagnostic mismatch between EmpkinS's SCID-5-CV labels and the E-DAIC threshold of PHQ-8 $\geq 10$), training on all EmpkinS participants (EmpkinS$_{\rm all} \to$ E-DAIC) achieves an AUC of 0.70 in the Preparatory/All configuration. 
Two compounding factors likely drive this performance: first, the passive Preparatory phase shares the highest functional similarity with the naturalistic listening portions of the E-DAIC interview; second, pooling all four conditions maximizes the available training data, providing sufficient variance to stabilize the classifier against domain shifts. When restricting the source data to the 80\% training split (EmpkinS $\to$ E-DAIC), forward transfer remains viable, peaking at an AUC of 0.65 for the active Neg./CR condition.

The reverse transfer direction (E-DAIC $\to$ EmpkinS) yields higher AUCs across multiple configurations. When evaluating this direction, it is important to note that all condition-specific test sets within EmpkinS are inherently small ($n_{\rm test} \leq 13$). Only the pooled Prep./All configuration aggregates a larger evaluation sample. Therefore, the exceptionally high AUCs observed in Setting 1, which reach 0.88 for Prep./AFE, 0.79 for Prep./SHAM, and even 1.00 for Pos./AFE, must be interpreted with caution. Although they clearly indicate a strong rank-predictive signal, the absolute point estimates remain volatile due to the restricted sample size. 
Similarly, in Setting 2, the peak performance of 0.85 for Neg./SHAM is subject to the exact same structural instability. Despite these high AUC values, the corresponding low F1$_{\rm M}$ scores in these specific cells (for instance, F1$_{\rm M} = 0.33$ for Pos./AFE) reveal the effect of domain shift on model calibration. Models trained on the highly imbalanced, low-severity E-DAIC dataset output conservative probability estimates when evaluating the structured RCT tasks, frequently classifying all participants into a single class. This preserves the rank-ordering measured by the AUC while failing at the standard 0.5 decision threshold measured by the F1 score. Despite this calibration shift, the consistent presence of a strong rank-predictive signal across both passive observation (Preparatory) and active emotion regulation phases indicates that models trained on semi-structured interviews can successfully detect binary depressive patterns across a diverse range of structured RCT contexts. 

To contextualize these transfer outcomes, we must examine the within-corpus baselines. The EmpkinS $\to$ EmpkinS baseline achieves an AUC of 0.78 (Neg./CR) in Setting 1 and reaches 0.91 (Neg./CR) in Setting 2. This high internal consistency confirms the presence of strong discriminative facial signals within the RCT environment. Conversely, the E-DAIC $\to$ E-DAIC baseline (which is configuration-independent) struggles, yielding an AUC of only 0.58. This low baseline performance underscores the inherent difficulty of detecting depression in a semi-structured setting with a heavily right-skewed severity distribution. Given this weak starting point, the ability of E-DAIC-trained models to achieve high AUCs when transferring to EmpkinS highlights the robustness of the underlying categorical biomarkers.

Finally, the divergence in transferability between classification and regression
highlights a fundamental property of facial biomarkers. The relative success of
binary classification suggests that categorical MDD status is encoded in global,
context-robust facial features, such as an overall reduction in expressivity,
pervasive psychomotor retardation, or baseline affective blunting~\cite{girard2014nonverbal,
cohn2009detecting}. Continuous severity estimation, conversely, requires
fine-grained, localized gradations of facial movement that remain tied to the
specific elicitation context. This context-dependence of continuous biomarkers
is the key mechanistic explanation for the asymmetric transfer outcomes observed
across tasks, and underscores why passive elicitation contexts, which leave
natural facial behavior undisturbed, are a necessary condition for
regression transfer.

\section{Conclusion}
We presented a cross-corpus study of facial biomarkers for depression
assessment, pairing a structured RCT with a naturalistic clinical interview corpus (E-DAIC). Continuous PHQ-8 regression transfer fails across most configurations in the forward direction, while binary MDD classification transfers more robustly, achieving an AUC of 0.70 in the Preparatory/All configuration. Reverse regression transfer succeeds modestly only during the passive Preparatory baseline phase; the one RCT context functionally most comparable to an unstructured interview. Notably, matched PHQ-8 labels (Setting~2) do not consistently outperform clinician-confirmed SCID-5-CV labels (Setting~1), suggesting that diagnostic label reliability can partially compensate for nominal label mismatch. 

Together, these results establish functional context alignment as the primary determinant of cross-corpus generalization in facial depression assessment, operating above and beyond shared feature extractors, label instruments, and training set size. Passive elicitation contexts produce the most transferable models, while structured active regulation phases yield strong within-corpus signal that does not survive domain shift. Future AI-based mental health assessment tools should therefore distinguish between deployment contexts: passive observation suffices for cross-context robustness, while structured affective tasks are preferable when the target domain is known and within-corpus sensitivity is prioritized. Evaluations should further distinguish between binary detection and continuous severity estimation, as these represent fundamentally different generalization challenges.

\section*{Acknowledgments}
We extend our gratitude to the students who supported the study execution and data collection. We also thank the software development team for their technical contributions to the EmpkinS-EKSpression application. This study is funded by the German Research Foundation (Deutsche Forschungsgemeinschaft, DFG) - SFB 1483 – Project-ID 442419336, EmpkinS.

\bibliographystyle{ACM-Reference-Format}
\bibliography{sample-sigconf}

@String{Computing = "Computing" }

@String{Computer = "{IEEE} Computer" }

@String{Springer = "Springer-Verlag" }

@ArtifactSoftware{R,
    title = {R: A Language and Environment for Statistical Computing},
    author = {{R Core Team}},
    organization = {R Foundation for Statistical Computing},
    address = {Vienna, Austria},
    year = {2019},
    url = {https://www.R-project.org/},
}

@misc{keinert2024facing,
  title={Facing depression: evaluating the efficacy of the EmpkinS-EKSpression reappraisal training augmented with facial expressions--protocol of a randomized controlled trial},
  author={Keinert, Marie and Schindler-Gmelch, Lena and Rupp, Lydia Helene and Sadeghi, Misha and Capito, Klara and Hager, Malin and Rahimi, Farnaz and Richer, Robert and Egger, Bernhard and Eskofier, Bjoern M and others},
  journal={BMC psychiatry},
  volume={24},
  number={1},
  pages={896},
  year={2024},
  publisher={Springer}
}

@book{beck1979cognitive,
  title={Cognitive Therapy of Depression},
  author={Beck, A. T. and Rush, J. and Shaw, B. F. and Emery, G.},
  year={1979},
  publisher={Guilford Press},
  address={New York, NY, USA},
  pages={442}
}

@article{kliem2018bhs,
  title={Psychometric properties and measurement invariance of the Beck Hopelessness Scale (BHS): results from a German representative population sample},
  author={Kliem, S. and Lohmann, A. and M{\"o}{\ss}le, T. and Br{\"a}hler, E.},
  journal={BMC Psychiatry},
  volume={18},
  number={1},
  pages={110},
  year={2018}
}

@article{possel2005automatische,
  title={Evaluation eines deutschsprachigen Instrumentes zur Erfassung positiver und negativer automatischer Gedanken},
  author={P{\"o}ssel, P. and Seemann, S. and Hautzinger, M.},
  journal={Z. Klin. Psychol. Psychother.},
  volume={34},
  number={1},
  pages={27--34},
  year={2005}
}

@article{ferring1996rosenberg,
  title={Messung des Selbstwertgef{\"u}hls: Befunde zur Reliabilit{\"a}t, Validit{\"a}t, und Stabilit{\"a}t der Rosenberg-Skala [Measurement of self-esteem: findings on reliability, validity, and stability of the Rosenberg Scale]},
  author={Ferring, D. and Filipp, S. H.},
  journal={Diagnostica},
  volume={42},
  pages={284--292},
  year={1996}
}

@inproceedings{gratch2014distress,
  title={The distress analysis interview corpus of human and computer interviews},
  author={Gratch, Jonathan and Artstein, Ron and Lucas, Gale M and Stratou, Giota and Scherer, Stefan and Nazarian, Angela and Wood, Rachel and Boberg, Jill and DeVault, David and Marsella, Stacy and others},
  booktitle={LREC},
  pages={3123--3128},
  year={2014},
  organization={Reykjavik}
}

@inproceedings{devault2014simsensei,
  title={SimSensei Kiosk: A virtual human interviewer for healthcare decision support},
  author={DeVault, David and Artstein, Ron and Benn, Grace and Dey, Teresa and Fast, Ed and Gainer, Alesia and Georgila, Kallirroi and Gratch, Jon and Hartholt, Arno and Lhommet, Margaux and others},
  booktitle={Proceedings of the 2014 international conference on Autonomous agents and multi-agent systems},
  pages={1061--1068},
  year={2014}
}

@inproceedings{ringeval2019avec,
  title={AVEC 2019 workshop and challenge: state-of-mind, detecting depression with AI, and cross-cultural affect recognition},
  author={Ringeval, Fabien and Schuller, Bj{\"o}rn and Valstar, Michel and Cummins, Nicholas and Cowie, Roddy and Tavabi, Leili and Schmitt, Maximilian and Alisamir, Sina and Amiriparian, Shahin and Messner, Eva-Maria and others},
  booktitle={Proceedings of the 9th International on Audio/visual Emotion Challenge and Workshop},
  pages={3--12},
  year={2019}
}

@inproceedings{baltruvsaitis2016openface,
  title={Openface: an open source facial behavior analysis toolkit},
  author={Baltru{\v{s}}aitis, Tadas and Robinson, Peter and Morency, Louis-Philippe},
  booktitle={2016 IEEE winter conference on applications of computer vision (WACV)},
  pages={1--10},
  year={2016},
  organization={IEEE}
}

@inproceedings{chen2016xgboost,
  author       = {Chen, Tianqi and Guestrin, Carlos},
  title        = {{XGBoost}: A Scalable Tree Boosting System},
  booktitle    = {Proceedings of the 22nd ACM SIGKDD International Conference on
                  Knowledge Discovery and Data Mining},
  pages        = {785--794},
  year         = {2016},
  publisher    = {ACM},
  doi          = {10.1145/2939672.2939785}
}

@article{lin1989ccc,
  author       = {Lin, Lawrence I-Kuei},
  title        = {A Concordance Correlation Coefficient to Evaluate Reproducibility},
  journal      = {Biometrics},
  volume       = {45},
  number       = {1},
  pages        = {255--268},
  year         = {1989},
  doi          = {10.2307/2532051}
}

@article{sadeghi2024harnessing,
  title={Harnessing multimodal approaches for depression detection using large language models and facial expressions},
  author={Sadeghi, Misha and Richer, Robert and Egger, Bernhard and Schindler-Gmelch, Lena and Rupp, Lydia Helene and Rahimi, Farnaz and Berking, Matthias and Eskofier, Bjoern M},
  journal={npj Mental Health Research},
  volume={3},
  number={1},
  pages={66},
  year={2024},
  publisher={Nature Publishing Group UK London}
}

@inproceedings{sadeghi2023exploring,
  title={Exploring the capabilities of a language model-only approach for depression detection in text data},
  author={Sadeghi, Misha and Egger, Bernhard and Agahi, Reza and Richer, Robert and Capito, Klara and Rupp, Lydia Helene and Schindler-Gmelch, Lena and Berking, Matthias and Eskofier, Bjoern M},
  booktitle={2023 IEEE EMBS International Conference on Biomedical and Health Informatics (BHI)},
  pages={1--5},
  year={2023},
  organization={IEEE}
}

@inproceedings{sadeghi2026crosscorpus,
  author    = {Sadeghi, Misha and Nguyen, Phuc Truong Loc and
               Triantafyllopoulos, Andreas and Habibpour, Mahdis and
               Richer, Robert and Rupp, Lydia Helene and
               Schindler-Gmelch, Lena and Keinert, Marie and
               Hager, Malin and Schuller, Bj{\"o}rn W. and
               Egger, Bernhard and Berking, Matthias and
               Eskofier, Bjoern M.},
  title     = {Cross-Corpus Depression Detection in Semi-Structured Clinical Interviews},
  booktitle = {Proceedings of the 2026 Conference on Empirical Methods in Natural Language Processing},
  year      = {2026},
  month     = oct,
}

@article{rupp2025stress,
  author       = {Rupp, Lydia Helene and Kumar, Akash and Sadeghi, Misha and
                  Schindler-Gmelch, Lena and Keinert, Marie and Eskofier, Bjoern M.
                  and Berking, Matthias},
  title        = {Stress Can Be Detected During Emotion-Evoking Smartphone Use:
                  A Pilot Study Using Machine Learning},
  journal      = {Frontiers in Digital Health},
  volume       = {7},
  year         = {2025},
  doi          = {10.3389/fdgth.2025.1578917}
}

@techreport{who2023,
  author       = {{World Health Organization}},
  title        = {Depressive Disorder (Depression): Fact Sheet},
  institution  = {World Health Organization},
  year         = {2023},
  note         = {\url{www.who.int/news-room/fact-sheets/detail/depression}}
}

@article{cummins2015review,
  author       = {Cummins, Nicholas and Scherer, Stefan and Krajewski, Jarek and
                  Schnieder, Sebastian and Epps, Julien and Quatieri, Thomas F.},
  title        = {A Review of Depression and Suicide Risk Assessment Using Speech
                  Analysis},
  journal      = {Speech Communication},
  volume       = {71},
  pages        = {10--49},
  year         = {2015},
  doi          = {10.1016/j.specom.2015.03.004}
}

@inproceedings{valstar2016avec2016,
  author       = {Valstar, Michel and Gratch, Jonathan and Schuller, Bj\"{o}rn and
                  Ringeval, Fabien and Lalanne, Denis and Torres Torres, Mercedes and
                  Scherer, Stefan and Stratou, Giota and Cowie, Roddy and
                  Pantic, Maja},
  title        = {{AVEC} 2016: Depression, Mood, and Emotion Recognition Workshop
                  and Challenge},
  booktitle    = {Proceedings of the 6th International Workshop on Audio/Visual
                  Emotion Challenge (AVEC)},
  pages        = {3--10},
  year         = {2016},
  publisher    = {ACM},
  doi          = {10.1145/2988257.2988258}
}

@inproceedings{jain2014depression,
  author       = {Jain, Shashank and Bhatt, Preethi and Bhatt, Neeraj and
                  Nair, Sreeja S.},
  title        = {Depression Analysis Using Action Unit and Eye Gaze Features},
  booktitle    = {Proceedings of the 4th International Workshop on Audio/Visual
                  Emotion Challenge (AVEC)},
  pages        = {41--48},
  year         = {2014},
  publisher    = {ACM},
  doi          = {10.1145/2661806.2661815}
}

@inproceedings{SchullerRingeval2015,
  author       = {Schuller, Bj\"{o}rn and Ringeval, Fabien and Marchi, Erik and
                  Deng, Jiankun and Lalanne, Denis and Valstar, Michel and Cowie, Roddy},
  title        = {{AVEC} 2015: The 5th International Audio/Visual Emotion Challenge
                  and Workshop},
  booktitle    = {Proceedings of the 2015 ACM International Conference on Multimedia},
  pages        = {1335--1336},
  year         = {2015},
  publisher    = {ACM},
  doi          = {10.1145/2733373.2807408}
}

@article{pampouchidou2017automatic,
  author       = {Pampouchidou, Anastasia and Simos, Panagiotis G. and Marias, Kostas
                  and Meriaudeau, Fabrice and Yang, Fan and Pediaditis, Matthew and
                  Tsiknakis, Manolis},
  title        = {Automatic Assessment of Depression Based on Visual Cues: A
                  Systematic Review},
  journal      = {{IEEE} Transactions on Affective Computing},
  volume       = {10},
  number       = {4},
  pages        = {445--470},
  year         = {2017},
  doi          = {10.1109/TAFFC.2017.2724035}
}

@inproceedings{yang2017dcnn,
  author       = {Yang, Le and Jiang, Dongmei and He, Lang and Peng, Ercheng and
                  Oveneke, Meshia C\'{e}dric and Sahli, Hichem},
  title        = {Decision Tree Based Depression Classification from Audio Visual
                  and Language Features},
  booktitle    = {Proceedings of the 7th Annual Workshop on Audio/Visual Emotion
                  Challenge (AVEC)},
  pages        = {89--96},
  year         = {2017},
  publisher    = {ACM},
  doi          = {10.1145/3133944.3133948}
}

@article{zhu2017multimodal,
  author       = {Zhu, Yongwei and Shang, Yu and Shao, Zhenwei and Guo, Guodong},
  title        = {Automated Depression Diagnosis Based on Deep Networks to Encode
                  Facial Appearance and Dynamics},
  journal      = {{IEEE} Transactions on Affective Computing},
  volume       = {9},
  number       = {4},
  pages        = {578--584},
  year         = {2017},
  doi          = {10.1109/TAFFC.2017.2650899}
}

@article{matsumoto2008cultural,
  author       = {Matsumoto, David and Yoo, Seung Hee and Nakagawa, Satoko},
  title        = {Culture, Emotion Regulation, and Adjustment},
  journal      = {Journal of Personality and Social Psychology},
  volume       = {94},
  number       = {6},
  pages        = {925--937},
  year         = {2008},
  doi          = {10.1037/0022-3514.94.6.925}
}

@article{alghowinem2016multimodal,
  author       = {Al-Ghowinem, Sharifa and Goecke, Roland and Wagner, Michael and
                  Epps, Julien and Parker, Gordon and Breakspear, Michael},
  title        = {Multimodal Depression Detection: Fusion Analysis of Paralinguistic,
                  Head Pose and Eye Gaze Behaviors},
  journal      = {{IEEE} Transactions on Affective Computing},
  volume       = {9},
  number       = {4},
  pages        = {478--490},
  year         = {2016},
  doi          = {10.1109/TAFFC.2016.2634527}
}

@inproceedings{baltrusaitis2018openface,
  author    = {Baltru{\v{s}}aitis, Tadas and Zadeh, Amir and Lim, Yao Chong and Morency, Louis-Philippe},
  title     = {OpenFace 2.0: Facial Behavior Analysis Toolkit},
  booktitle = {2018 13th IEEE International Conference on Automatic Face \& Gesture Recognition (FG 2018)},
  year      = {2018},
  pages     = {59--66},
  doi       = {10.1109/FG.2018.00019}
}

@book{ekman1978,
  author       = {Ekman, Paul and Friesen, Wallace V.},
  title        = {Facial Action Coding System: A Technique for the Measurement of
                  Facial Movement},
  publisher    = {Consulting Psychologists Press},
  address      = {Palo Alto, CA},
  year         = {1978}
}

@article{ke2017lightgbm,
  title={Lightgbm: A highly efficient gradient boosting decision tree},
  author={Ke, Guolin and Meng, Qi and Finley, Thomas and Wang, Taifeng and Chen, Wei and Ma, Weidong and Ye, Qiwei and Liu, Tie-Yan},
  journal={Advances in neural information processing systems},
  volume={30},
  year={2017}
}

@article{prokhorenkova2018catboost,
  title={CatBoost: unbiased boosting with categorical features},
  author={Prokhorenkova, Liudmila and Gusev, Gleb and Vorobev, Aleksandr and Dorogush, Anna Veronika and Gulin, Andrey},
  journal={Advances in neural information processing systems},
  volume={31},
  year={2018}
}

@article{cuijpers2010self,
  title={Self-reported versus clinician-rated symptoms of depression as outcome measures in psychotherapy research on depression: a meta-analysis},
  author={Cuijpers, Pim and Li, Juan and Hofmann, Stefan G and Andersson, Gerhard},
  journal={Clinical psychology review},
  volume={30},
  number={6},
  pages={768--778},
  year={2010},
  publisher={Elsevier}
}

@article{dong2025foundation,
  title={Foundation Model-Based Evaluation of Neuropsychiatric Disorders: A Lifespan-Inclusive, Multi-Modal, and Multi-Lingual Study},
  author={Dong, Zhongren and Guo, Haotian and Xu, Weixiang and Zhao, Huan and Zhang, Zixing},
  journal={IEEE Journal of Selected Topics in Signal Processing},
  year={2025},
  publisher={IEEE}
}

@inproceedings{cohn2009detecting,
  title={Detecting depression from facial actions and vocal prosody},
  author={Cohn, Jeffrey F and Kruez, Tomas Simon and Matthews, Iain and Yang, Ying and Nguyen, Minh Hoai and Padilla, Margara Tejera and Zhou, Feng and De la Torre, Fernando},
  booktitle={2009 3rd international conference on affective computing and intelligent interaction and workshops},
  pages={1--7},
  year={2009},
  organization={IEEE}
}

@article{girard2014nonverbal,
  title={Nonverbal social withdrawal in depression: Evidence from manual and automatic analyses},
  author={Girard, Jeffrey M and Cohn, Jeffrey F and Mahoor, Mohammad H and Mavadati, S Mohammad and Hammal, Zakia and Rosenwald, Dean P},
  journal={Image and vision computing},
  volume={32},
  number={10},
  pages={641--647},
  year={2014},
  publisher={Elsevier}
}

@INPROCEEDINGS{alghowinem2016cross,
  author={Alghowinem, Sharifa and Goecke, Roland and Cohn, Jeffrey F. and Wagner, Michael and Parker, Gordon and Breakspear, Michael},
  booktitle={2015 11th IEEE International Conference and Workshops on Automatic Face and Gesture Recognition (FG)}, 
  title={Cross-cultural detection of depression from nonverbal behaviour}, 
  year={2015},
  volume={1},
  number={},
  pages={1-8},
  doi={10.1109/FG.2015.7163113}}

\clearpage

\appendix
\onecolumn

\section{Feature Selection and Model Hyperparameters}
\label{app:hyperparams}

\subsection{Feature representation}
Each OpenFace signal column is summarized into 18 statistical functionals:
mean, std, min, max, skew, kurtosis, range, and Shannon entropy applied to
both the original signal and its first-order frame differences, plus
rate-of-change and peak count on the original signal.
Applied to 49 raw OpenFace columns (17 AU intensities, 17 AU presence
indicators, 6 gaze, 4 head pose rotations, 4 head pose translations, 1
confidence column), this yields the 882-feature shared set used in all
cross-corpus experiments.

\subsection{Feature selection}

\paragraph{Step 1: Univariate filter}
\emph{Regression}: features are ranked by absolute Spearman $|\rho|$; those
surviving Benjamini--Hochberg FDR correction ($q = 0.20$) are retained, with a
fallback to the top 20 by $|\rho|$ when fewer than 20 pass the threshold.
\emph{Classification}: Mann--Whitney $U$ test; same BH-FDR procedure.

\paragraph{Step 2: RFE}
Recursive feature elimination (Table~\ref{tab:reg_grids},
Table~\ref{tab:cls_grids}) with a Ridge (regression) or Logistic Regression
(\texttt{lbfgs}, \texttt{max\_iter = 2000}, classification) surrogate;
$k \in \{5, 10, 15, 20\}$ candidate sizes evaluated by 3-fold CV (MAE for
regression, AUC for classification); the $k$ minimizing the CV score is
selected as the final feature count. Across experiments, the selected $k$
typically falls in the range 5--20 features, with smaller per-condition
subsets frequently selecting $k = 5$ or $k = 10$.

\subsection{Regression models and hyperparameter grids}
Hyperparameter grids (Table~\ref{tab:reg_grids}) are searched with 5-fold
\texttt{GridSearchCV} scored by negative MAE. Three scalers (Standard,
MinMax, Robust) are evaluated jointly with each model; the best
scaler--model combination is selected.

\begin{table}[H]
\centering
\small
\caption{Regression hyperparameter search grids. All tree ensembles use
\texttt{random\_state=42}; MLP uses \texttt{max\_iter=500}.}
\label{tab:reg_grids}
\begin{tabular}{ll}
\toprule
\textbf{Model} & \textbf{Search grid} \\
\midrule
Ridge        & \texttt{alpha} $\in \{0.01,\,0.1,\,1,\,10,\,100\}$ \\
Lasso        & \texttt{alpha} $\in \{0.01,\,0.1,\,1,\,10\}$;
               \texttt{max\_iter=5000} \\
ElasticNet   & \texttt{alpha} $\in \{0.01,\,0.1,\,1\}$,\;
               \texttt{l1\_ratio} $\in \{0.2,\,0.5,\,0.8\}$;
               \texttt{max\_iter=5000} \\
SVR (RBF)    & \texttt{C} $\in \{0.1,\,1,\,10,\,100\}$,\;
               \texttt{gamma} $\in \{\texttt{scale},\,\texttt{auto}\}$,\;
               \texttt{epsilon} $\in \{0.1,\,0.5,\,1.0\}$ \\
$k$-NN       & \texttt{n\_neighbors} $\in \{3,\,5,\,7,\,9\}$ \\
Random Forest & \texttt{n\_estimators} $\in \{100,\,200\}$,\;
               \texttt{max\_features} $\in \{\sqrt{\cdot},\,0.3\}$,\;
               \texttt{min\_samples\_leaf} $\in \{1,\,3\}$ \\
Extra Trees  & \texttt{n\_estimators} $\in \{100,\,200\}$,\;
               \texttt{max\_features} $\in \{\sqrt{\cdot},\,0.3\}$ \\
Gradient Boosting & \texttt{n\_estimators} $\in \{100,\,200\}$,\;
               \texttt{learning\_rate} $\in \{0.05,\,0.1\}$,\;
               \texttt{max\_depth} $\in \{3,\,5\}$ \\
XGBoost      & \texttt{n\_estimators} $\in \{100,\,200\}$,\;
               \texttt{learning\_rate} $\in \{0.05,\,0.1\}$,\;
               \texttt{max\_depth} $\in \{3,\,5\}$;
               \texttt{eval\_metric=mae} \\
LightGBM     & \texttt{n\_estimators} $\in \{100,\,200\}$,\;
               \texttt{learning\_rate} $\in \{0.05,\,0.1\}$,\;
               \texttt{num\_leaves} $\in \{15,\,31\}$ \\
CatBoost     & \texttt{iterations} $\in \{100,\,200\}$,\;
               \texttt{learning\_rate} $\in \{0.05,\,0.1\}$,\;
               \texttt{depth} $\in \{4,\,6\}$ \\
MLP          & \texttt{hidden\_layer\_sizes} $\in \{(64),\,(128),\,(64{,}32)\}$,\;
               \texttt{alpha} $\in \{10^{-4},\,10^{-3}\}$ \\
\bottomrule
\end{tabular}
\end{table}
\subsection{Classification models and hyperparameter grids}
Hyperparameter grids (Table~\ref{tab:cls_grids}) are searched with 5-fold
\texttt{GridSearchCV} scored by AUC (same three scalers as regression).

\begin{table}[H]
\centering
\small
\caption{Classification hyperparameter search grids. All tree ensembles use
\texttt{random\_state=42}; MLP uses \texttt{max\_iter=500};
Logistic Regression uses \texttt{solver=lbfgs}, \texttt{max\_iter=2000}.}
\label{tab:cls_grids}
\begin{tabular}{ll}
\toprule
\textbf{Model} & \textbf{Search grid} \\
\midrule
Logistic Regression & \texttt{C} $\in \{0.01,\,0.1,\,1,\,10\}$ \\
SVC (RBF)      & \texttt{C} $\in \{0.1,\,1,\,10\}$,\;
                 \texttt{gamma} $\in \{\texttt{scale},\,\texttt{auto}\}$ \\
Random Forest  & \texttt{n\_estimators} $\in \{100,\,200\}$,\;
                 \texttt{max\_features} $\in \{\sqrt{\cdot},\,0.3\}$,\;
                 \texttt{min\_samples\_leaf} $\in \{1,\,3\}$ \\
Extra Trees    & \texttt{n\_estimators} $\in \{100,\,200\}$,\;
                 \texttt{max\_features} $\in \{\sqrt{\cdot},\,0.3\}$ \\
Gradient Boosting & \texttt{n\_estimators} $\in \{100,\,200\}$,\;
                 \texttt{learning\_rate} $\in \{0.05,\,0.1\}$,\;
                 \texttt{max\_depth} $\in \{3,\,5\}$ \\
XGBoost        & \texttt{n\_estimators} $\in \{100,\,200\}$,\;
                 \texttt{learning\_rate} $\in \{0.05,\,0.1\}$,\;
                 \texttt{max\_depth} $\in \{3,\,5\}$ \\
LightGBM       & \texttt{n\_estimators} $\in \{100,\,200\}$,\;
                 \texttt{learning\_rate} $\in \{0.05,\,0.1\}$,\;
                 \texttt{num\_leaves} $\in \{15,\,31\}$ \\
CatBoost       & \texttt{iterations} $\in \{100,\,200\}$,\;
                 \texttt{learning\_rate} $\in \{0.05,\,0.1\}$,\;
                 \texttt{depth} $\in \{4,\,6\}$ \\
MLP            & \texttt{hidden\_layer\_sizes} $\in \{(64),\,(128),\,(64{,}32)\}$,\;
                 \texttt{alpha} $\in \{10^{-4},\,10^{-3}\}$ \\
\bottomrule
\end{tabular}
\end{table}

\section{SHAP Feature Importance}
\label{app:shap}

\begin{figure*}[t]
\centering
\includegraphics[width=12cm]{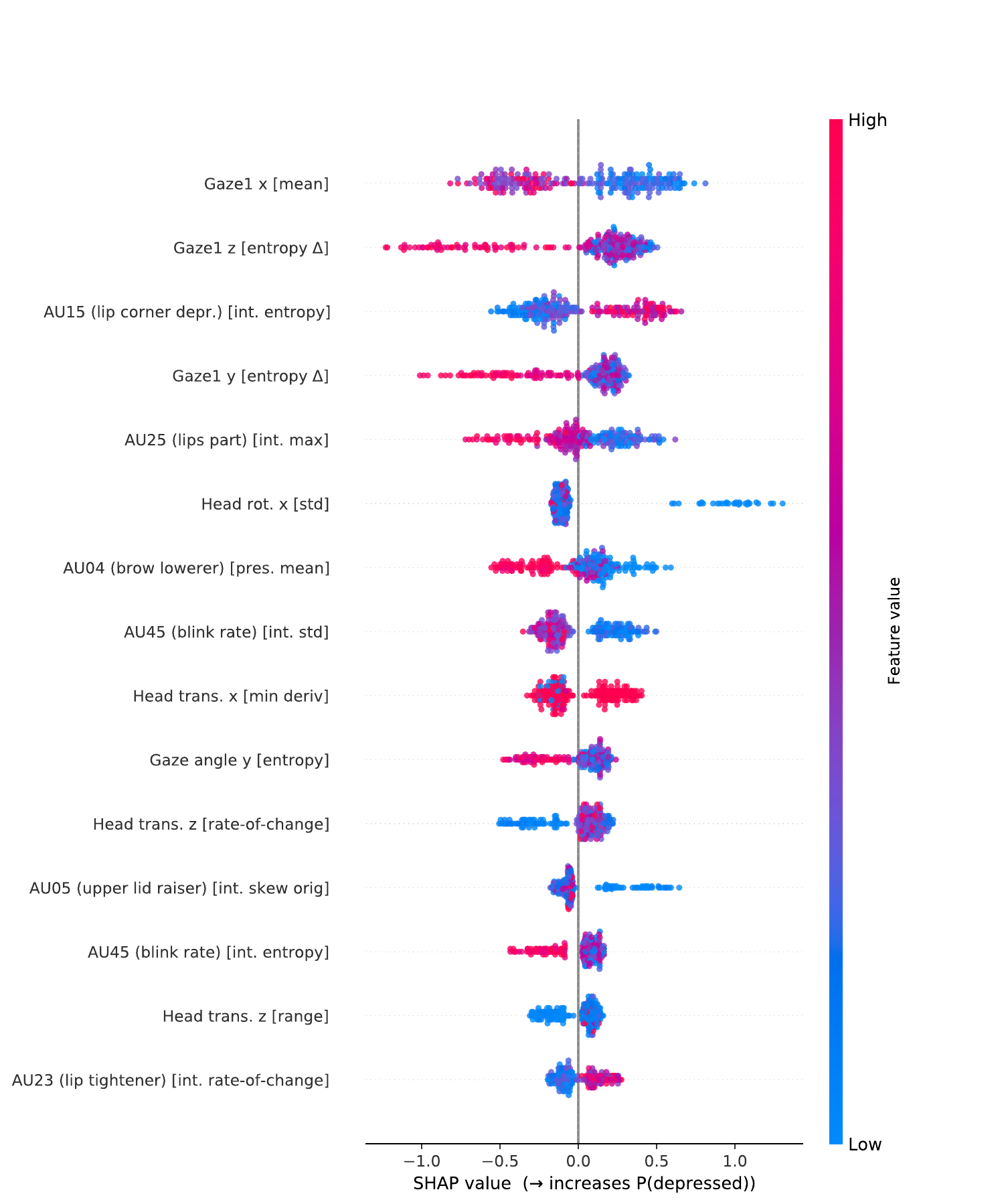}
\caption{SHAP beeswarm plot for the best forward-transfer classifier:
EmpkinS-EKSpression (EmpkinS)\,$\to$\,E-DAIC (Prep./All, Setting~1, AUC\,=\,0.70). Features
ranked by mean $|$SHAP$|$; computed on the source training partition
to reflect learned feature importance.}
\label{fig:shap1}
\end{figure*}

\begin{figure*}[t]
\centering
\includegraphics[width=12cm]{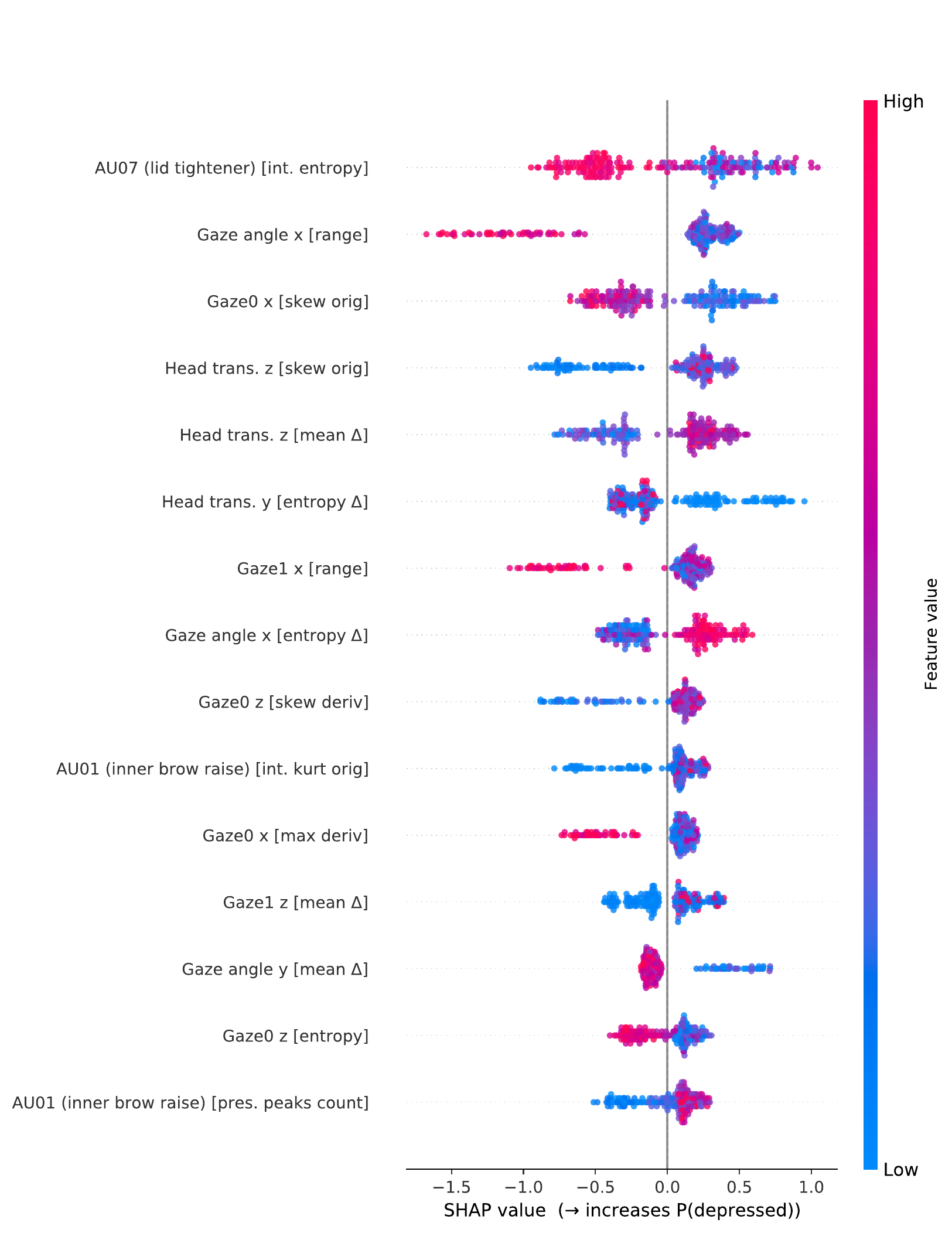}
\caption{SHAP beeswarm plot for the best reverse-transfer classifier:
E-DAIC\,$\to$\,EmpkinS-EKSpression (EmpkinS) (Pos./AFE, Setting~1, AUC\,=\,1.00). Features ranked by mean $|$SHAP$|$; computed on the source training partition to reflect learned feature importance.}
\label{fig:shap2}
\end{figure*}

Figures~\ref{fig:shap1} and~\ref{fig:shap2} show SHAP beeswarm plots for
the two best-performing cross-corpus classifiers, computed using a LightGBM
interpretability model trained on the source partition. Each point represents
one participant; color indicates feature value (red = high, blue = low);
horizontal position indicates the feature's impact on the model output.

In the forward direction (EmpkinS\,$\to$\,E-DAIC, Fig.~\ref{fig:shap1}),
gaze dynamics dominate (Gaze1 $x$ mean, Gaze1 $z$/$y$ entropy of frame
differences), followed by AU15 (lip-corner depressor) intensity entropy,
head rotation variability, AU04 (brow lowerer), and AU25 (lips part).
These features align with established depression markers of restricted gaze
range, blunted lip activity, and reduced head movement.

In the reverse direction (E-DAIC\,$\to$\,EmpkinS, Fig.~\ref{fig:shap2}),
gaze features dominate even more strongly (gaze angle range, gaze direction
skewness and entropy), alongside AU07 (lid tightener) intensity entropy,
head translation dynamics, and AU01 (inner brow raise). The predominance
of gaze-based features in both directions suggests that gaze restriction
is the most cross-corpus-stable facial marker of depression in this feature
set.

\end{document}